\documentclass[10pt,twocolumn]{article}
\usepackage{amssymb}
\usepackage{amsmath}
\usepackage{float}
\usepackage[colorlinks=true]{hyperref}
\usepackage{times}
\usepackage{epsfig}
\usepackage{graphicx}
\usepackage[acronym]{glossaries}
\glsdisablehyper
\usepackage{tocloft}
\usepackage{subcaption}
\usepackage{authblk}

\usepackage{multirow}
\usepackage{booktabs}
\newacronym{cnn}{CNN}{convolution-neural-network}
\newacronym{vit}{ViT}{vision-transformer}
\newacronym{mlp}{MLP}{multi-layer perceptron}
\newacronym{mse}{MSE}{mean-squared error}
\newacronym{mae}{MAE}{mean-absolute error}
\newacronym{sed}{SED}{symmetric epipolar distance}
\newacronym{std}{STD}{standard deviation}
\newacronym{fsl}{FSL}{few-shot learning}

\title{A Comparative Study of Vision Transformers and CNNs for Few-Shot Rigid Transformation and Fundamental Matrix Estimation}

\begin{document}

\author[1]{Alon Kaya\thanks{alonkay@post.bgu.ac.il}}
\author[2]{Igal Bilik}
\author[3]{Inna Stainvas}

\affil[1]{Department of Computer Science, Ben-Gurion University of the Negev, Beer Sheva, Israel}
\affil[2]{School of Electrical and Computer Engineering, Ben-Gurion University of the Negev, Beer Sheva, Israel}
\affil[3]{GE Healthcare Science and Technology, Israel}

\date{}  % Removes the date

\maketitle

\begin{abstract}
\Glspl{vit}  and large-scale \glspl{cnn} have reshaped computer vision through pretrained feature representations that enable strong transfer learning for diverse tasks. However, their efficiency as backbone architectures for geometric estimation tasks involving image deformations in low-data regimes remains an open question. 
This work considers two such tasks: 1) estimating 2D rigid transformations between pairs of images and 2) predicting the fundamental matrix for stereo image pairs, an important problem in various applications, such as autonomous mobility, robotics, and 3D scene reconstruction.
Addressing this intriguing question, this work systematically compares large-scale \glspl{cnn} (ResNet, EfficientNet, CLIP-ResNet) with \gls{vit}-based foundation models (CLIP-ViT variants and DINO) in various data size settings, including few-shot scenarios.
These pretrained models are optimized for classification or contrastive learning, encouraging them to focus mostly on high-level semantics. The considered tasks require balancing local and global features differently, challenging the straightforward adoption of these models as the backbone.
Empirical comparative analysis shows that, similar to training from scratch, \glspl{vit} outperform \glspl{cnn} during refinement in large downstream-data scenarios. However, in small data scenarios, the inductive bias and smaller capacity of \glspl{cnn} improve their performance, allowing them to match that of a \gls{vit}. Moreover, \glspl{vit} exhibit stronger generalization in cross-domain evaluation where the data distribution changes.
These results emphasize the importance of carefully selecting model architectures for refinement, motivating future research towards hybrid architectures that balance local and global representations.
\end{abstract}

% %%Graphical abstract
% \begin{graphicalabstract}
% \includegraphics[width=\linewidth]{Small_err.png}
% \includegraphics[width=\linewidth]{Network.drawio.png}
% \end{graphicalabstract}

% \begin{highlights}
% \item  A comparative investigation of transfer learning \glspl{vit} and \glspl{cnn} for geometric transformation estimations as regression tasks with varying dataset sizes.
% \item Even with a deep pretrained \gls{vit}, the \gls{cnn} achieves comparable or even superior performance in the low-data regime during transfer learning.
%  \item The fundamental matrix can be accurately estimated in a few-shot setting, outperforming other state-of-the-art approaches with comparable training data size.
% \item A proposed network that enables a fair comparison between diverse models for various image deformation estimation tasks.
% \end{highlights}

% %% Keywords
% \begin{keyword}
% Vision-Transformers \sep Epipolar-Geometry \sep Few-Shot \sep Regression-Tasks \sep Transfer-Learning 
% \end{keyword}

% \end{frontmatter}

% \linenumbers

\glsreset{vit}
\glsreset{cnn}
\section{Introduction}
In recent years, large-scale models, such as \glspl{cnn} and \glspl{vit}~\cite{animageisworth}, have revolutionized the computer vision domain. When trained using large datasets, \glspl{cnn} and \glspl{vit} enable efficient feature transfer with minimal fine-tuning. They were shown to provide exceptional generalization capabilities and can be efficiently used as frozen or refined backbones for diverse downstream tasks~\cite{animageisworth, howtotrainyour}.  

Recently, these models were extensively compared in the literature, mostly when trained from scratch~\cite {tokens2token} or fine-tuned for classification problems~\cite{understanding_robustness_of}. It was demonstrated that given sufficient pre-training data, \glspl{vit} outperform \glspl{cnn} when fine-tuned for classification tasks~\cite{animageisworth}. 
In addition, differences in feature extraction mechanisms between these models have been investigated, and it was shown that larger \glspl{vit} enable more robust intermediate representations, thus providing robustness in classification tasks~\cite{do_vision_transformers},~\cite{intriguing_preoperties},~\cite{vision_transformers_are}. 
It was also shown that for segmentation tasks, \gls{cnn}-based models outperform \glspl{vit} when trained from scratch on smaller datasets, \cite{A_Comparative_Analysis_of_Vision}. The efficiency of transfer learning from general image classification to more specialized recognition scenarios was demonstrated in ~\cite{alonso2024combined}. It was shown that the off-the-shelf features extracted from both pretrained \glspl{cnn} and \glspl{vit} are suitable for periocular recognition.

The limited availability of annotated training data is one of the major practical challenges~\cite{Intra_task_for_Few_Shot}, and the \gls{fsl} framework was introduced to address it~\cite{Few_Shot_Classification}. \Gls{fsl} has been used in various tasks, such as natural language understanding~\cite{Language_Models_are_Few_Shot}, image classification~\cite{LIU2024110640}, semantic segmentation~\cite{fsll}, object detection~\cite{Meta_DETR}, and others~\cite{HAN2025111326}.

This work focuses on \gls{cnn} and \gls{vit} comparison by applying transfer learning to downstream regression tasks involving geometric estimations of image distortions while investigating the influence of training data size, including a minimal few-shot learning framework.
Unlike conventional high-level semantic tasks, such as image classification, segmentation, and object detection, geometric transformation tasks require both more intensive local feature extraction (e.g., for implicit pixel correspondences) and global scene understanding (e.g., scene geometry)~\cite{YANG2025110876}. These requirements challenge the \glspl{vit} and \glspl{cnn} adoptions as backbone architectures for the considered geometric estimation tasks. 
Two downstream tasks are considered in this work: (i) estimating 2D rigid transformations between pairs of images and (ii) predicting the fundamental matrix, $\mathbf{F}$-matrix, between a pair of stereo images in an end-to-end manner.
We utilize these tasks across different models to conduct a comparative analysis between \glspl{vit} and \glspl{cnn}, addressing the following related open questions:
\begin{itemize}
\item How does the parameter-count difference between \glspl{vit} and \glspl{cnn} affect refinement performance in few-shot settings? 
\item How do differences in pretraining objectives influence performance in these downstream tasks?
\end{itemize}
The \gls{vit} and \gls{cnn} are extensively tested and compared with the state-of-the-art models. It is demonstrated that: 
\begin{itemize}
\item Transfer learning using \glspl{vit} and \glspl{cnn} enables efficient feature extraction for accurately estimating geometric transformations, even in a challenging few-shot scenario.
\item \glspl{vit} outperform \glspl{cnn} during refinement in larger-data scenarios, while \glspl{cnn} achieve comparable performance in low-data regimes even when the \glspl{vit} were pretrained on large-scaled datasets.
\item Smaller patch-size \glspl{vit} outperform their larger patch-size variants in geometric transformation tasks.
\item Freezing bottom layers in the \gls{vit} can enhance its performance during refinement by reducing overfitting in small-data scenarios.
\item All considered models generalize well to an unseen dataset for the fundamental matrix estimation task, and the \glspl{vit} outperform \glspl{cnn} in terms of cross-domain dataset generalization.
\end{itemize}

The main novelty of this work can be summarized as follows:
\begin{itemize}
    \item  A comparative investigation of transfer learning \glspl{vit} and \glspl{cnn} for geometric transformation estimations as regression tasks with varying dataset sizes.
    \item The finding that, even with a deep pretrained \gls{vit}, the \gls{cnn} achieves comparable or even superior performance in the low-data few-shot regime during transfer learning.
     \item It is demonstrated that the fundamental matrix can be accurately estimated between stereo image pairs in a few-shot setting, outperforming other state-of-the-art approaches with comparable training data size.
    \item A proposed network that enables a fair comparison between diverse models for various image deformation estimation tasks.
\end{itemize}
The results of this work are expected to contribute to a broad spectrum of applications, such as aerial photography, medical imaging, and autonomous vehicles. 
    
\section{Background and Related Work}\label{RW}

\subsection{ViT vs. CNN-based Models}

Transformers, originally introduced for natural language processing (NLP)~\cite{attention_is}, were adapted to computer vision~\cite{animageisworth} by converting images into a sequence of fixed-size patches that are linearly projected into embeddings, and include a global classification (CLS) token.

Comparison between \glspl{vit} and \gls{cnn}-based models has been extensively investigated in the literature~\cite{animageisworth} \cite{tokens2token}. \Glspl{cnn} were shown to outperform conventional \gls{vit} architectures when trained from scratch or when pretrained using limited data~\cite{animageisworth} \cite{tokens2token}. This superiority can be attributed to \glspl{cnn}' inductive bias towards locality by focusing on local connections in the lower layers and weight sharing. \glspl{vit} commonly have larger capacities and, therefore, require significantly larger data to estimate self-attention parameters accurately~\cite{animageisworth}. When sufficient data is available, \glspl{vit} efficiently capture global dependencies and commonly outperform \glspl{cnn}. 

It was also shown that largely-pretrained \glspl{vit} are robust to occlusions and domain shifts and provide higher generalization capabilities using off-the-shelf features for classification tasks~\cite{intriguing_preoperties} \cite{vision_transformers_are}.
Furthermore, it was demonstrated that smaller patch sizes in \glspl{vit} enhance their robustness to adversarial spatial transformations~\cite{understanding_robustness_of}. It was also shown that, unlike \glspl{cnn}, which discard low-level details to achieve discriminative representation in higher layers, \glspl{vit} retain detailed image information across all layers, providing strong feature propagation~\cite{visualization_comparison_of_vit}.
Further, it was shown that spatial information, which is processed in the early layers, is crucial to achieving high performance and that during transfer learning, the lower layers of a \gls{vit} remain well preserved in classification tasks~
\cite{do_vision_transformers}. The effect of pretraining dataset size on transfer learning performance was investigated, revealing that \glspl{vit} provide stronger intermediate representations through larger pretraining datasets~\cite{do_vision_transformers}.
However, comparative studies between the \gls{vit} and \gls{cnn}-based models with transfer learning for geometric transformation estimations while analyzing downstream data size have not been conducted in the literature.

\subsection{The Fundamental Matrix Estimation Problem}
Estimation of the fundamental matrix, $\mathbf{F}$, is one of the key problems in computer vision~\cite{deepf_no_cors}. The $\mathbf{F}$-matrix describes the intrinsic projective relationships between two pin-hole cameras observing the same 3D scene from different viewpoints. For any matching point pair with homogeneous coordinates, $\mathbf{p} = [x,y,1]^T$, in the first, and $\mathbf{q} = [x', y', 1]^T$, in the second image, the following epipolar constraint has to be satisfied:
\begin{equation}
 \mathbf{q}^T \mathbf{F} \mathbf{p} = 0\;,  \label{eq:epi_constraint} 
\end{equation}
meaning the match in the second image lies on an epipolar line determined by: $\ell_e = \mathbf{F} \mathbf{p}$.
The fundamental matrix, $\mathbf{F}$, is defined by the camera intrinsic, $\mathbf{K}_1$ and $\mathbf{K}_2$, and the relative pose between cameras, encapsulated by rotation, $\mathbf{R}$, and translation, $\mathbf{t}$ as: 
\begin{equation} 
\mathbf{F} = \mathbf{K}_2^{-T} [\mathbf{t}]_\times R \mathbf{K}_1^{-1}\;, \label{eq: F_decompose} 
\end{equation}
where $\mathbf{F}$ is a $3 \times 3$ rank-$2$ matrix that contains the essential geometry between two views. The fundamental matrix facilitates outlier detection in point correspondences by constraining their search to epipolar lines, enhancing stereo matching and 3D reconstruction algorithms~\cite{Fundamental_Matrix_of_a_Stereo_Pair}.\\

\noindent \textbf{Estimating the F-matrix with deep learning}.\\
Conventional $\mathbf{F}$-matrix estimation approaches, such as SIFT~\cite{sift} and RANSAC~\cite{ransac}, focus on sparse feature matching and robust estimation. Recently, deep learning-based feature matching approaches like D2Net~\cite{d2_net}, SuperPoint~\cite {superpoint}, and R2D2~\cite{r2d2} have also been proposed. Alternative approaches, such as RoMa~\cite{roma} and DKM~\cite{dkm}, exploit dense feature matching by pixel-wise correspondences, improving accuracy in texture-less regions.
The $\mathbf{F}$-matrix estimation problem is closely related to estimating the pose matrix, which has also been explored using deep learning approaches~\cite{ Regression_Based_Camera_Pose_Est} \cite{yossikeller}.

Alternative approaches attempt to estimate the $\mathbf{F}$-matrix directly in an end-to-end manner. The deep learning-based approach for the direct $\mathbf{F}$-matrix estimation was first introduced in~\cite{deepf}.
Siamese-built \gls{cnn}-based network and a rank constraint layer for the $\mathbf{F}$-matrix estimation without point correspondences was proposed in~\cite{deepf_no_cors}.
The approach for the fundamental matrix estimation from bi-planar radiographs using a specialized end-to-end network, enhancing estimation accuracy, was introduced in~\cite{FM-net}. 
An alternative end-to-end network that integrates feature correspondence with outlier rejection and employs a novel loss function using an evaluation metric more suitable for visual tasks was investigated in~\cite{an_end_to_end}. 
A twin-feature pyramid architecture with an improved attention module was proposed to estimate the fundamental matrix directly from images without correspondences~\cite{estimating_the_f}. 
This work follows the direct $\mathbf{F}$-matrix estimation approach, but unlike previous work, it focuses on F-estimation in a data-limited regime.

\section{Methodology}\label{sec:na}
This work investigates the performance of pretrained \gls{vit} and \gls{cnn} models for geometric estimation tasks, considering data-limited scenarios, including a few-shot learning framework with only 32 samples. The performance is evaluated for two geometric estimation tasks: 2D rigid transformation and $\mathbf{F}$-matrix estimation. The first task requires predicting three parameters of 2D geometric transformations: the rotation angle around the image center and the horizontal and vertical translations. 
The $\mathbf{F}$-matrix estimation task involves complex 3D geometric transformations between pairs of stereo images.

\begin{figure*}[t]
    \centering
    \includegraphics[width=1\linewidth]{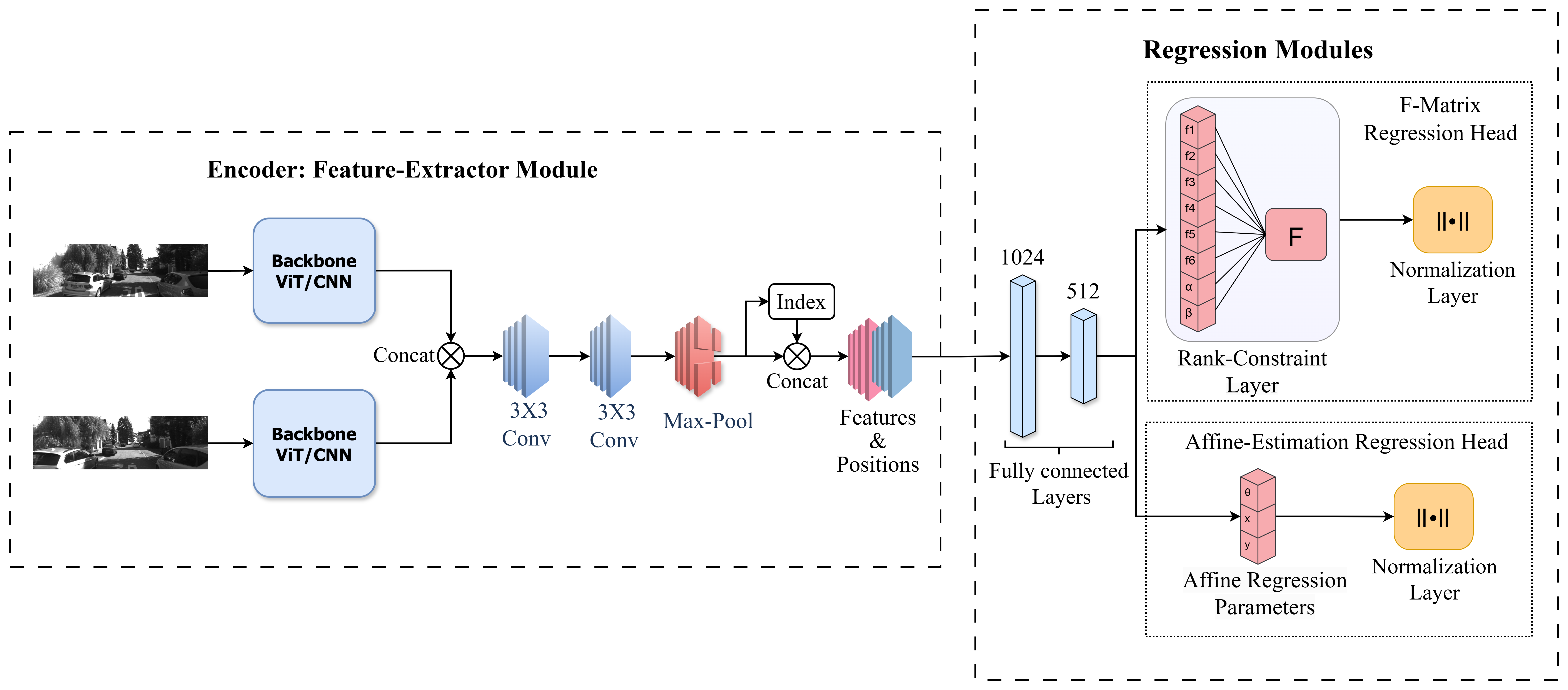}
    \caption{Schematic representation of the proposed network for comparison between \glspl{vit} and \glspl{cnn} used as the backbone models. The network contains encoder and regression modules. The \gls{vit} outputs patch embeddings and an unused CLS token for each frame. The \gls{cnn} outputs aggregated feature maps. The backbone outputs of both frames are reshaped to a data structure of the size, $(D, \sqrt{N}, \sqrt{N})$, where $D$ is the embedding dimension and $N$ is the number of spatial patches for \glspl{vit} or number of pixels in feature maps for \glspl{cnn}. These embeddings are concatenated along the depth dimension and passed through two convolutional layers, followed by location-aware Max-pooling. 
 The regression module for the $\mathbf{F}$-matrix estimation task consists of a rank-constraint layer, and the rigid estimation consists of the $3$ transformation elements, each using different normalization methods.}
    \label{fig:network}
\end{figure*}

This work introduces a dedicated network architecture, shown in Fig.~\ref{fig:network}, that enables the fair comparison between the \gls{vit} and \gls{cnn}-based models. The proposed network consists of two modules: 1) an encoder for feature extraction and 2) a regression for task-specific predictions. The encoder module is kept the same for all considered tasks, and the regression head is designed specifically for each task as detailed in Subsections~\ref{sec:regression_affine} and~\ref{sec:regression_F}. The pretrained \gls{vit} or \gls{cnn}-based models were used as the encoder backbone. 

We consider structurally varied CNN and ViT models, differing in parameter count, pre-training objectives, and training datasets, summarized in Table~\ref{tab:models}.
The \glspl{vit} represent state-of-the-art large-scale foundation models trained using large datasets for various tasks. The considered CLIP-VIT models~\cite{clip} represent vision language models (VLMs) that are trained for text-image alignment with different patch sizes. The \gls{vit} with a smaller patch size captures finer image details, providing better image-text alignment performance. The DINO-ViT-B/16 vision model~\cite{dino} relies on contrastive self-distilled learning principles. It is trained using the ImageNet dataset~\cite{imagenet}, which is significantly smaller than the size of the data used for CLIP training but shares a similar \gls{vit} architecture. 
These ViT-based models are compared to CNN-based baselines, including ResNet-152~\cite{resnet} and 
EfficientNetV2-M~\cite{Efficient}, both trained for image classification on ImageNet, and have fewer parameters than the ViT models. We also evaluate CLIP-ResNet-101 and CLIP-ResNet-50x4~\cite{clip}, which share the same training objectives as the CLIP-ViTs but use convolutional architectures. Notably, CLIP-ResNet-50x4 has a number of parameters comparable to the ViT-based models, making it particularly relevant for model-size comparisons.
\begin{table}[ht]
\centering
\scriptsize  
\setlength{\tabcolsep}{4pt} % Reduce space between columns (default is 6pt)
\caption{The considered backbone models. ImageNet1k consists of $1.2$M images, and the WebImageText dataset contains $400$M image-text pairs collected from the internet. ViT-B/32 and ViT-B/16 denote \gls{vit} models with patch size, $32$ and $16$, respectively, and SSL denotes the self-supervised learning.}
\begin{tabular}{|c|c|c|c|} 
\hline 
\textbf{Model} & \textbf{\# Params} & \textbf{Dataset}  &\textbf{Training Task}\\ 
\hline  
\multicolumn{4}{|c|}{\textbf{Convolutional Networks (CNNs)}} \\  
\hline
ResNet-152 \cite{resnet} & 60M & ImageNet1k & Classification \\  
EfficientNetV2-M \cite{Efficient} &  54.1M & ImageNet1k & Classification \\  
CLIP-ResNet-101 \cite{clip} & 56.4M & WebImageText  & Image–Text Alignment \\
CLIP-ResNet-50X4 \cite{clip} & 87.2M & WebImageText  & Image–Text Alignment \\
\hline  
\multicolumn{4}{|c|}{\textbf{Vision Transformers (ViTs)}} \\  
\hline
CLIP-ViT-B/32 \cite{clip} & 88.2M & WebImageText  & Image–Text Alignment \\  
CLIP-ViT-B/16 \cite{clip} & 86.5M & WebImageText  & Image–Text Alignment \\  
DINO-ViT-B/16 \cite{dino} & 85.8M & ImageNet1k & SSL \\  
\hline
\end{tabular}
\label{tab:models}
\end{table}

% \begin{singlespace}
% \end{singlespace}

\subsection{Encoder: Feature Extraction Module}
The feature extraction module, optimized via the ablation study detailed in Section~\ref{sec:short_ablation}, extracts meaningful features from two input images independently. Each image is processed separately using a pretrained model as a Siamese-like network with shared weights. The output of \gls*{vit} models consists of a series of embedding tokens that correspond to features extracted from image patches and a global CLS token, which is ignored. Each token can interact with every other token from the initial layers of the \gls{vit}, creating a global receptive field. Differently, the \glspl{cnn} produce aggregated features across increasingly larger receptive fields, with each layer capturing larger spatial relationships. 

The feature vectors of the two images are then reshaped to a data structure with dimensions: $(D,\sqrt{N},\sqrt{N})$, where $D$ is the embedding dimension and $N$ is the number of spatial patches for \glspl{vit} or number of pixels in feature maps for \glspl{cnn}. These vectors are concatenated depth-wise, forming a unified representation. This representation is processed by two $3\times 3$ convolutional layers with output channel sizes of $256$ and $512$, respectively. Each layer is followed by ReLU activation and batch normalization to ensure stable and efficient training.

\subsubsection{Location-Aware Max-pooling} 
The considered tasks rely on preserving the spatial relationships of corresponding image points. Conventional down-sampling operations like Max-Pooling tend to lose this critical location information. Therefore, this work uses the location-aware max-pooling layer~\cite{deepf_no_cors} after the convolutional layers. This layer selects the most significant features while retaining their spatial locations by concatenating the max-pool output features with their (normalized) indices. This approach enables the network to maintain spatial information essential for accurate estimations of the considered image deformations.  

\section{Rigid Transformation Estimation}\label{sec:affine}
For planar scenes, rigid transforms are used to simulate a fronto-parallel stereo camera setup. In such cases, the geometric parameters which define the cameras’ relative pose correspond directly to those estimated in the fundamental matrix (F) computation. The idea was to first refine model performance on rigid transform estimation, where abundant synthetic data is available, and then use these refined models to initialize F-estimation. This could have reduced the dataset size needed and alleviated small-data limitations. However, this strategy did not produce improvements, so we did not pursue it further. Instead, this work focuses on comparing architectures directly in the small-data regime.

\subsection{Regression Module}\label{sec:regression_affine}
The proposed network in Fig.~\ref{fig:network} contains separate regression and encoder modules. The encoder outputs feature vectors with their spatial indices. This output is flattened and then processed by a \gls{mlp} with two fully connected layers with feature dimensions of $512$ and $1024$, each using a ReLU activation function. The \gls{mlp} output is a $3$-dimensional vector consisting of the estimated rotation angle and horizontal and vertical shifts. The output values are normalized to the range $[-1,1]$.

\subsection{Training and Evaluation Setup}
The network hyperparameters were optimized using the Adam optimizer, selecting the best combination of learning rates from $6\times10^{-5}$ and $1\times10^{-4}$ and batch sizes from $32$ and $4$.

\subsubsection{Loss Function}
The proposed loss function combines the \gls{mse} and Huber losses, detailed in \ref{sec:A}, as follows:  
\begin{equation}
{\cal L} = {\cal L}^{\theta}_{MSE}  + {\cal L}^{\theta}_{Huber} + \alpha  ({\cal L}^{s}_{MSE}  + {\cal L}^{s}_{Huber})\;,
\label{eq:affine_loss}
\end{equation}
where ${\cal L}^{\theta}_{MSE}$ and ${\cal L}^{\theta}_{Huber}$ are the \gls{mse} and Huber loss for angle estimations, respectively, and ${\cal L}^{s}_{MSE}$ and ${\cal L}^{s}_{Huber}$ are the corresponding shift estimation losses. The hyperparameter \(\alpha\) was empirically set to $10$, balancing the angle and the shift contributions to the total loss for estimation accuracy optimization.

\subsubsection{Performance Evaluation Metrics}
The proposed comparative analysis is performed using translation and rotation error metrics. Translation error is computed as the Euclidean, $L_2$, distance, measured in pixels, between the predicted shifts along the $x$ and $y$ axes and their corresponding ground-truth values. Rotation error is evaluated using the \gls{mae}, in degrees, between the predicted rotation angle and the ground-truth angle. The Euclidean distance and \gls{mae} metrics formulas are given in \ref{sec:A}.

\subsection{Dataset}
The considered dataset is a subset of ImageNet~\cite{frgfm_imagenette}, selected for its smaller scale and suitability to our low-data regime. Since the task involves estimating rigid transformations rather than semantic classification, the full ImageNet dataset was unnecessary. The subset provides sufficient visual diversity while reducing computational and data overhead. The experiments were conducted using training data of varying sizes, with smaller sizes being subsets of the larger ones. Models were trained multiple times on disjoint sub-trajectories of the dataset, each with the same size, reporting the mean and \gls{std} of the performance across all subsets. Each image was resized to $256\times256$, followed by a random crop of size $224\times 224$. Other data augmentation techniques included brightness, contrast, and Gaussian blur adjustments. The rotation angles were randomly sampled from the range $[-30^o, 30^o]$, while the translations along both the $x$- and $y$-axes were sampled between $[-32, 32]$. These values are then normalized to the range of $[-1,1]$. 

\subsection{Performance Evaluation}
\begin{figure}[t]
    \centering

    \begin{minipage}[t]{\linewidth}
        \centering
        \includegraphics[width=\linewidth]{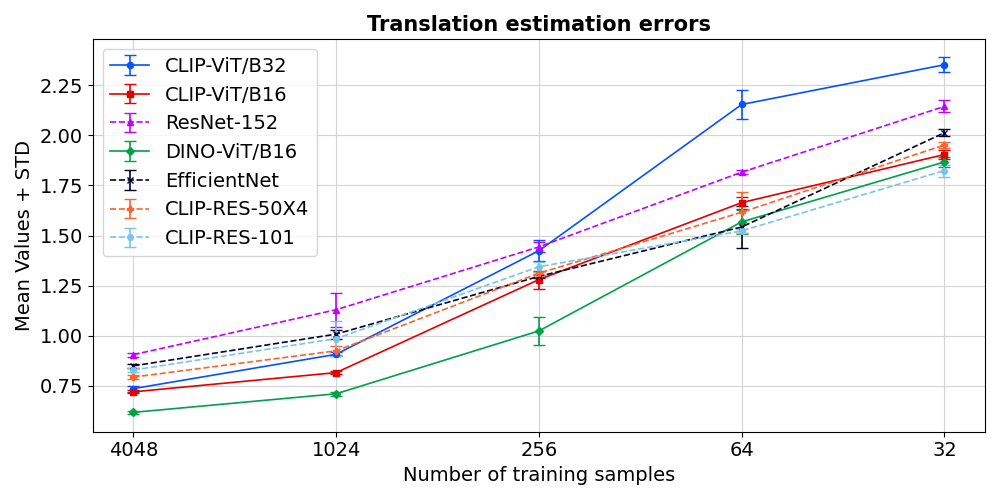}
        \textbf{(a)}
    \end{minipage}
    \begin{minipage}[t]{\linewidth}
        \centering
        \includegraphics[width=\linewidth]{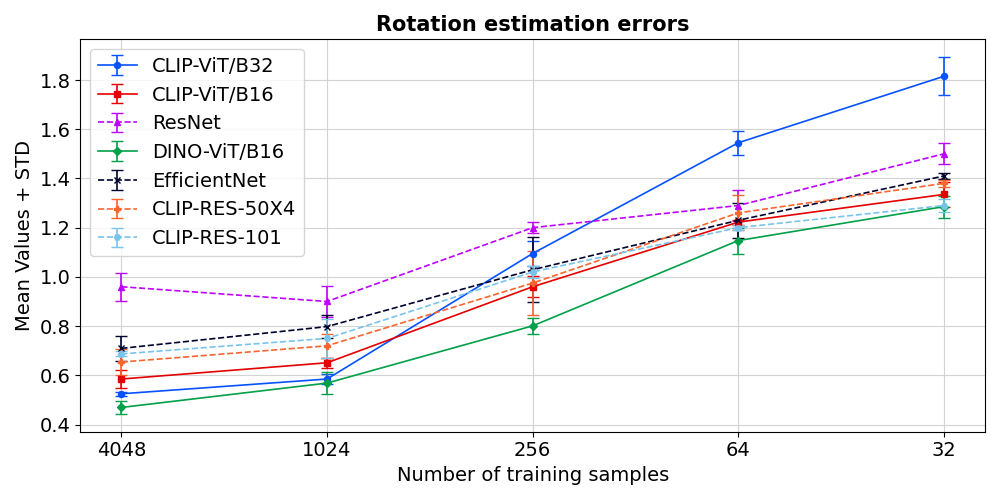}
        \textbf{(b)}
    \end{minipage}
    \caption{Rigid transformation estimation performance. (a) Euclidean translation error in pixels. (b) MAE rotation error in degrees. Solid lines represent ViT-based models and dashed lines represent CNN-based models. For the visualization, \glspl{std} are divided by factor, $3$.}
    \label{fig:affine}
\end{figure}

The rigid transformation estimation performance as a function of the amount of downstream training data is shown in Fig.~\ref {fig:affine}. 
Notice that the performance of all evaluated models degrades in scenarios with smaller available training data.
The \gls{vit} models outperform the \gls{cnn}-based models when trained using a large amount of data, even under identical training objectives and a comparable number of parameters, as demonstrated by the performance gap between the CLIP-ViT-B/16 and CLIP-Res-50X4 models. However, in the low-data regime, the performance of \glspl{vit} and \glspl{cnn} becomes comparable. The \gls{vit} with a smaller patch size of $16\times 16$ outperforms its larger patch-size variant in the majority of experiments. This confirms the claim of~\cite{understanding_robustness_of} that a \gls{vit} with a smaller patch size is more robust to spatial image distortions.     
Furthermore, although the DINO~\cite{dino} model is pretrained on a relatively smaller dataset, it observes a substantially larger number of image patches during self-supervised learning. This enables it to achieve performance comparable to, or even exceeding, that of CLIP-VIT~\cite{clip} in tasks such as rigid transformation estimation. This advantage stems from DINO’s ability to capture \textit{mid-level features with strong spatial locality}, which are particularly beneficial for tasks requiring precise geometric reasoning. Unlike CLIP, which is optimized for global semantic alignment across modalities and tends to lose spatial detail due to its pooling strategy~\cite{clipdinoiser2023}, DINO retains richer local structure through its self-distillation objective~\cite{densedino2023}. Recent studies have shown that DINO features exhibit higher sensitivity to patch-level variations and outperform CLIP in fine-grained perception tasks~\cite{frame2025}, making them more suitable for dense prediction and transformation estimation.
Among the evaluated models, CLIP-ResNet-101 achieves the best performance in the low-data regime. This can be attributed to its large-scale multimodal pretraining on hundreds of millions of image-text pairs, its relatively compact architecture compared to Transformer-based models, and the inductive bias introduced by convolutional layers. These factors contribute to its strong generalization ability when training data is limited. However, it is noteworthy that the DINO model achieves comparable performance, likely due to its ability to retain spatially localized mid-level features through self-supervised learning.

\section{Fundamental Matrix Estimation}\label{sec:F}
\subsection{Regression Module}\label{sec:regression_F}
The regression module for the fundamental matrix estimation in Fig.~\ref{fig:network} maps the extracted features and their spatial locations into a matrix, with the $\mathbf{F}$-matrix's properties, such as rank $2$ with $7$ degrees of freedom (DOF). 
The encoder module's output is flattened and processed by a \gls{mlp} with two fully connected layers with feature dimensions of $1024$ and $512$, using ReLU as the activation function. 
The \gls{mlp} produces an $8$-dimensional vector that serves as a compact representation of the matrix parameters. These parameters are then used to construct the $\mathbf{F}$-matrix in the subsequent rank-constrained layer~\cite{deepf_no_cors}. The first six elements of the vector form the first two columns, and the last two elements determine the third column as a linear combination of the first two.
Notice that the $\mathbf{F}$-matrix is scale-invariant. Therefore, a normalization layer is used to enforce this property and ensure numerical stability explicitly, $\mathbf{F}_{\text{normalized}}=\frac{\mathbf{F}}{|\mathbf{F}\|_{F_r}}$, where $\|\cdot\|_{F_r}$ is the matrix Frobenius norm\footnote{The ground truth fundamental matrix is normalized the same.}.

\subsection{Training and Evaluation Setup}
The network is trained using a learning rate of $1 \times 10^{-4}$ and a batch size of $8$, with the Adam optimizer. 
The \textit{\gls{sed}} symmetrically quantifies the geometric distances of corresponding points from their respective epipolar lines in the two images. The \gls{sed} is incorporated in the training loss to enforce epipolar consistency and is used during the test as an evaluation metric:
\begin{equation}
\text{SED}(\mathbf{F}, \mathbf{x}, \mathbf{x}') = \sum_{i=1}^n \left( 
 \frac{1}{\|l_i\|^2} +
 \frac{1}{\|l_i'\|^2}
\right)(x_i^T F x_i')^2\;,
\label{eq:sed}
\end{equation}
where $\mathbf{x}_i$ and $\mathbf{x}_i'$ are homogeneous corresponding points in the two images, and $n$ is the total number of point correspondences. The terms \(\|l'\|\) and \(\|l\|\) refer to the norms of the epipolar lines, $l'=\mathbf{F}^T \mathbf{x}$ and $l=\mathbf{F} \mathbf{x}'$, which normalize the epipolar distance, ensuring invariance to the image scale.

\subsubsection{Loss Function}
High-confidence inlier correspondence points with a \gls{sed} error below $0.01$ were selected using the ground-truth $\mathbf{F}$-matrix. These points were adjusted according to the resize and crop parameters and used to compute the \gls{sed} error component of the estimated $\mathbf{F}$-matrix, which contributes to the overall loss function:
\begin{equation}
{\cal L} = {\cal L}_{MSE} + \alpha {\cal L}_{Huber} + \beta {\cal L}_{SED}\;,
\label{eq:f_loss}
\end{equation}
where ${\cal L}_{MSE}$ and ${\cal L}_{Huber}$ are the MSE and Huber loss, calculated using the estimated and ground-truth $\mathbf{F}$-matrices and detailed in~\ref{sec:A}. The \( {\cal L}_{SED} \)  is the SED term  defined in~\eqref{eq:sed}, obtained from the estimated $\mathbf{F}$-matrix. The coefficients,  $\alpha$ and $\beta$, balance the contributions of the loss terms, and their optimal values are empirically set to $1$ and $10$, respectively.

\subsubsection{Performance Evaluation Metrics}
The fundamental matrix estimation accuracy is evaluated using two metrics: (i) SED, defined in~\eqref{eq:sed}, and (ii) the algebraic distance (AD), which evaluates the absolute error in satisfying the epipolar constraint~\eqref{eq:epi_constraint}:
\begin{equation}
{\text{AD}} = \sum_{i=1}^{n} |x_i^\top F x_i'|
\label{eq:alg}
\end{equation}
The two considered metrics demonstrated similar results in all conducted experiments. 

\subsection{Datasets}
Two data sets, KITTI Odometry dataset~\cite{KITTI} and FlyingThings3D dataset~\cite{sceneflow}, were considered in this work. The KITTI Odometry dataset is commonly used for real-world automotive applications and serves as a standard for testing vision algorithms in realistic settings. The FlyingThings3D dataset contains objects moving along randomized 3D trajectories without naturalistic constraints.

The performance was evaluated using training data of different sizes, obtained by sub-sampling each sequence with a different fraction, ensuring that data of smaller sizes are nested within larger ones. The dataset size ranges from $2166$ to $32$ image pairs; sizes below $32$ yield impractical SED values. Each sequence in the KITTI dataset was split into three equal, disjoint parts, reporting their mean and \gls{std}.
In the FlyingThings3D dataset, the training sequences were uniformly selected, while the test set remained fixed for both datasets.

Each image is resized to $256 \times 256$
 pixels and augmented using random cropping to $224\times224$
 pixels and adjustments to brightness, contrast, and Gaussian blur. 
Various ground-truth $\mathbf{F}$-matrices were generated and estimated based on the crop parameters.

\subsection{Performance Evaluation}
\subsubsection{Performance Evaluation Using KITTI Dataset}
\begin{figure}[t]
    \centering
    \begin{minipage}[t]{\linewidth}
        \centering
        \includegraphics[width=\linewidth]{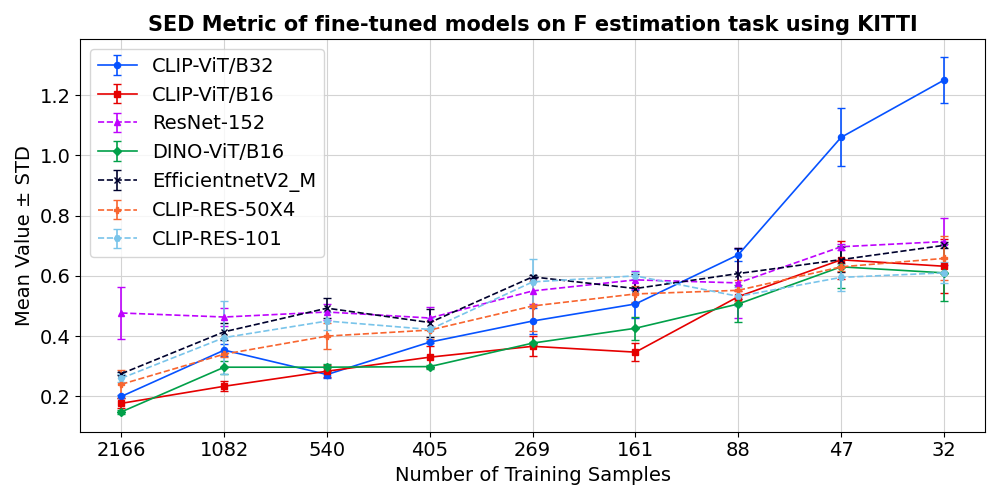}
        \textbf{(a)}
    \end{minipage}
    \begin{minipage}[t]{\linewidth}
        \centering
        \includegraphics[width=\linewidth]{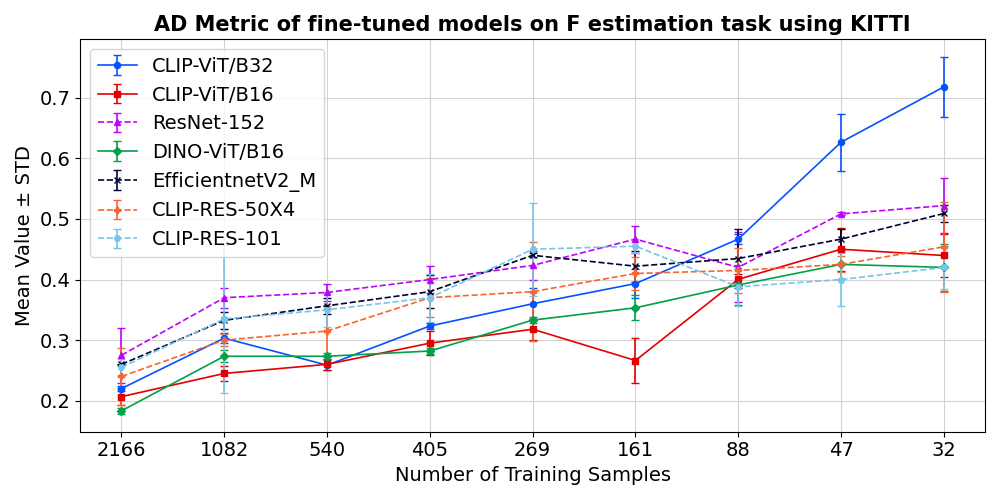}
        \textbf{(b)}
    \end{minipage}

    \caption{SED (a) and AD (b) metrics for $\mathbf{F}$-matrix estimation using \glspl{vit} and \gls{cnn} models with various training data sizes on the KITTI dataset. For visualization, \glspl{std} are divided by $3$. } 
    \label{fig:models_kitti}
\end{figure}

\begin{figure*}[t!]
    \centering
    \begin{subfigure}[b]{0.5\linewidth}
        \centering
         \includegraphics[width=\linewidth]{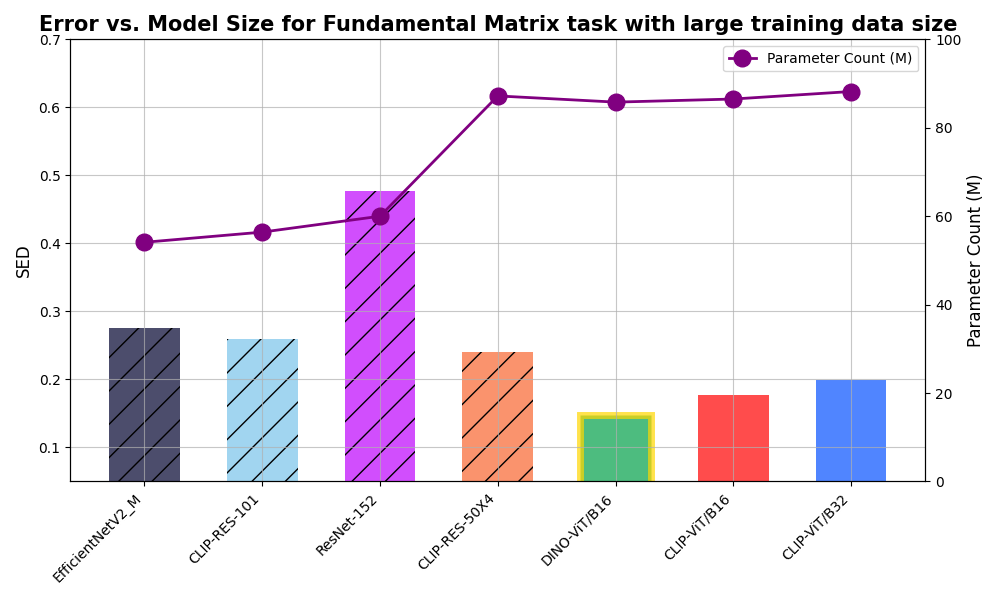}
        \textbf{(a)}
    \end{subfigure}%
    ~ 
    \begin{subfigure}[b]{0.5\linewidth}
        \centering
         \includegraphics[width=\linewidth]{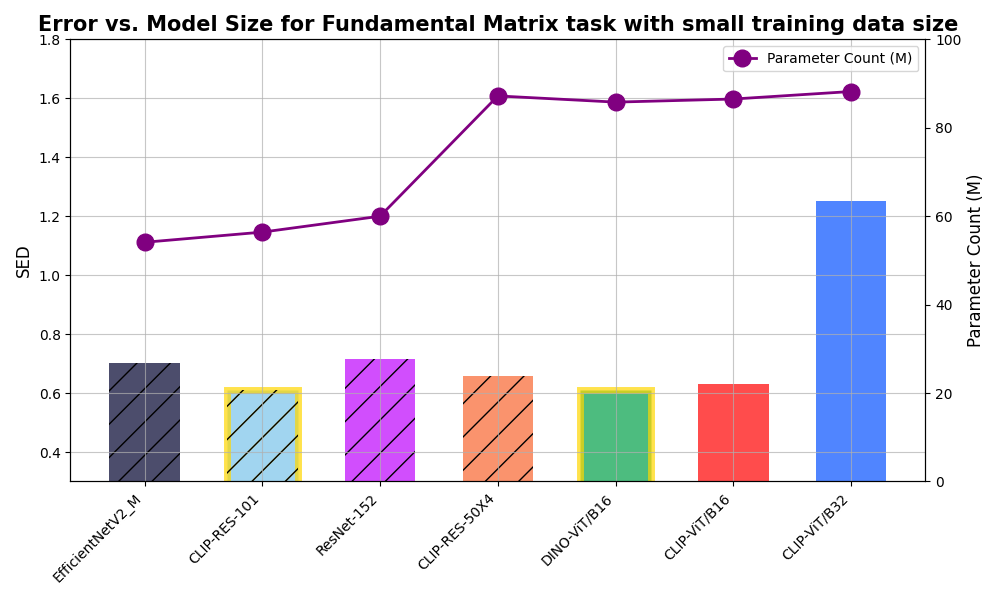}
        \textbf{(b)}
           \end{subfigure}

 \caption{SED results for the fundamental matrix estimation task on the KITTI dataset, comparing the largest (a) and smallest (b) training data sizes. Bar plots represent SED performance, and line plots indicate corresponding model parameter counts.
In the small-data regime (b), CLIP-ResNet-101 achieves the lowest SED, benefiting from extensive pretraining on image–text alignment using large-scale datasets, and a lower parameter count, which reduces overfitting. Among transformer-based models, DINO-ViT-B/16 slightly outperforms CLIP-ViT-B/16.
In the large-data regime (a), DINO-ViT-B16 achieves the best performance, outperforming all other models, including CLIP-ResNet variants. This highlights the advantage of its self-supervised learning (SSL) objective, particularly in tasks requiring fine-grained geometric reasoning.
Transformers with a patch size of 16 consistently perform well, likely due to better access to local spatial information. 
Overall, DINO-ViT-B16 emerges as the most stable model across varying data scales.
Our experiments also demonstrate that CLIP-ResNet-101 and CLIP-ResNet-50x4 perform best among CNN-based models, attributable to their training on large-scale WebImageText datasets using image–text alignment tasks. }
\label{fig:last}

\end{figure*}

\noindent{{\bf Training dataset size influence}}\\
Fig.~\ref{fig:models_kitti} shows the SED and AD of the $\mathbf{F}$-matrix estimation using \gls{vit} and \gls{cnn} models with various downstream training data sizes. 
Notice that in scenarios with large training data, \gls{vit}-based models consistently outperform  \glspl{cnn}, even under identical training objectives and a similar number of parameters. In scenarios with small data sizes, the \glspl{cnn} performance is comparable to the \glspl{vit} with a patch size of $16 \times 16$. \glspl{vit} with smaller patch sizes outperform the ones with larger patch sizes in the majority of tested scenarios due to their finer resolution needed for geometric and spatial transformation-related tasks. DINO-ViT-B/16~\cite{dino} and CLIP-ViT-B/16\cite{clip} achieve comparable results, although in the smallest data regime, the DINO model slightly outperforms CLIP-VIT once again, as a result of DINO's ability to retain strong local fine-grained features. The CLIP-RES-101 model achieves the best performance across the smallest data size, leveraging extensive pretraining on large-scale image–text alignment datasets and a lower number of parameters, which helps mitigate overfitting. According to the paired t-test~\cite{student1908probable}, the observed differences between CLIP-RES-101 and DINO-ViT/B16 are not statistically significant.
These conclusions are illustrated clearly in Fig.~\ref{fig:last} as well, which compares the models along with their sizes in the largest and smallest data regimes. These findings support the conclusions made in the rigid transformation estimation experiments.

The differences between the \gls{vit} and \gls{cnn}-based models' performance during refinement can be explained by the difference in the models' number of parameters and their architecture. This is consistent with results when these models are trained from scratch.
In a large data regime, the self-attention mechanism of the \glspl{vit} provides an advantage over \glspl{cnn}, even when compared to a similarly trained \gls{cnn} of a similar size. However, with limited data, the \glspl{cnn}' architecture, with its strong inductive bias towards locality, enables efficient extraction of spatial features in the lower layers~\cite{do_vision_transformers} \cite{animageisworth}. In contrast, even with a deep pretrained \gls{vit}, fine-tuning with limited data can be challenging, as \glspl{vit} rely heavily on global representations, which may be less effective at adapting to new tasks with small datasets. As a result, the advantage of \glspl{vit} diminishes under limited data, ultimately leading to performance comparable to that of \glspl{cnn}.

\noindent\textbf{Decreasing ViT Capacity During Refinement}\\
Beyond the architectural difference between \glspl{cnn} and \glspl{vit}, differences in the number of model parameters can also affect performance. The larger \gls{vit} models are expected to outperform the smaller \glspl{cnn} in large-data scenarios. However, networks with fewer parameters are less likely to overfit~\cite{overfit2}. Therefore, models with fewer parameters could be advantageous in small-data scenarios. Figs.~\ref{fig:affine}, \ref{fig:models_kitti}, and \ref{fig:last} show that the larger CLIP-ResNet model outperforms its smaller variant at larger data sizes but underperforms when training data is limited. 

\begin{figure}[t]
    \centering
    \begin{minipage}[t]{\linewidth}
        \centering
        \includegraphics[width=\linewidth]{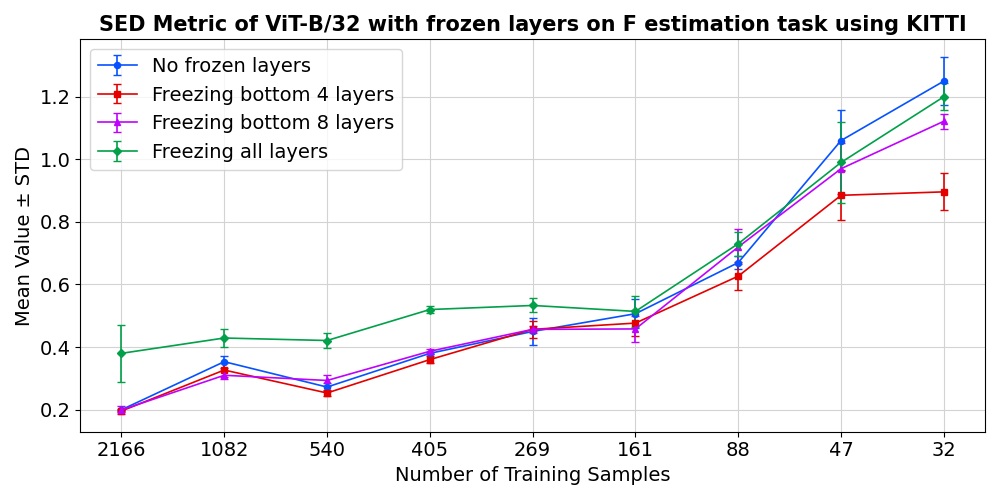}
        \textbf{(a)}
    \end{minipage}
    \begin{minipage}[t]{\linewidth}
        \centering
        \includegraphics[width=\linewidth]{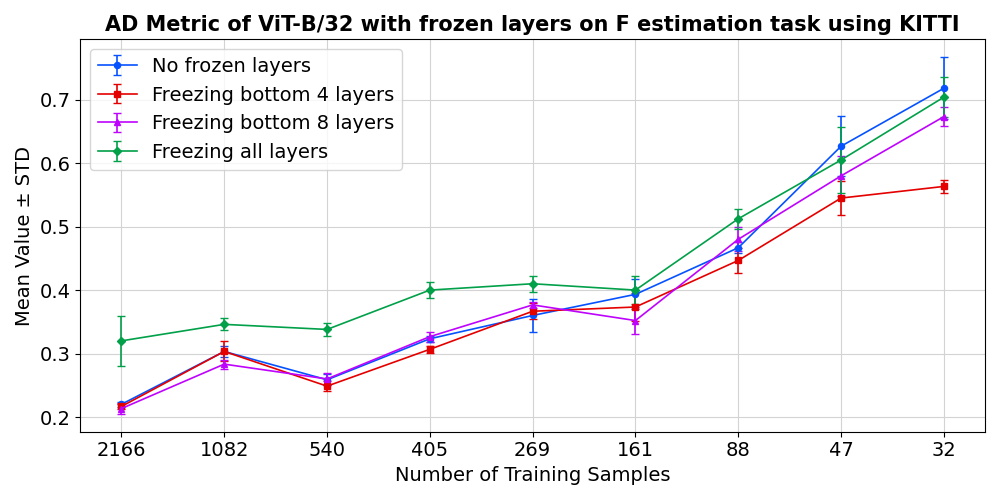}
        \textbf{(b)}
    \end{minipage}
        \caption{SED (a) and AD (b) for $\mathbf{F}$-matrix estimation using CLIP-ViT-B/32 freezing bottom layers. For visualization, \glspl{std} are divided by $3$.}
    \label{fig:frozen_low}
\end{figure}

This observation is further explored in the following experiments investigating the influence of the \gls{vit} capacity on the $\mathbf{F}$-matrix estimation performance for various training data sizes. 
Fig.~\ref{fig:frozen_low} shows the $\mathbf{F}$-matrix estimation errors of the CLIP-ViT-B/32 network, whose shallower layers were incrementally frozen to reduce the number of parameters down to a fully frozen network. Notice that in scenarios with extensive training data, fine-tuning some or all \gls{vit} layers yields similar errors, outperforming a fully frozen network. 
This can be explained by the \gls{vit}'s capability to perform task-specific adaptation when sufficient training data is available, outperforming a fixed-feature extractor.
In~\cite{do_vision_transformers}, it was suggested that \glspl{vit} capture hierarchical features, with bottom layers incorporating more low-level, task-agnostic features, which are essential for strong performance and remain largely preserved in transfer learning. Conversely, the deeper layers encode more global task-specific information. This observation explains the results in Fig.~\ref{fig:frozen_low}, where freezing bottom layers does not degrade performance even in large data regimes. 
Fig.~\ref{fig:frozen_low} shows that models with partially frozen layers outperform a fully trained network during refinement in data-limited scenarios. 
This can be explained by the observation that in data-limited scenarios, the bottom layers learn noise, and as a result, freezing these layers mitigates overfitting and stabilizes training. On the other hand, freezing the deeper layers reduces the model parameters. It limits the model's ability to capture task-specific features and adapt to downstream tasks, leading to degraded performance of the fully frozen network.
These results also support the relative performance improvement of the \glspl{cnn} with fewer parameters in data-limited scenarios during refinement.

% FIC INNA:
\subsubsection{Performance Evaluation Using FlyingThings3D Dataset}
\begin{figure}[t]
    \centering
    \begin{minipage}[t]{\linewidth}
        \centering
        \includegraphics[width=\linewidth]{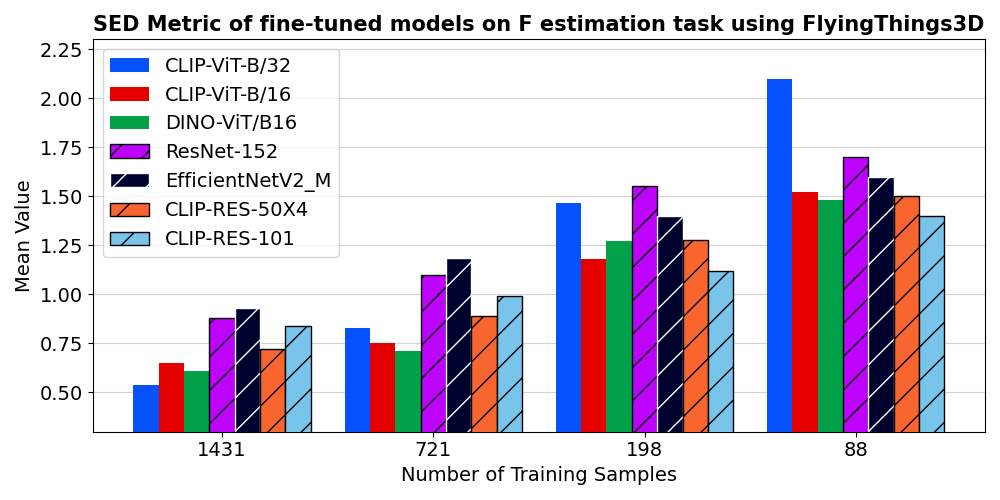}
        \textbf{(a)}
    \end{minipage}
    \begin{minipage}[t]{\linewidth}
        \centering
        \includegraphics[width=\linewidth]{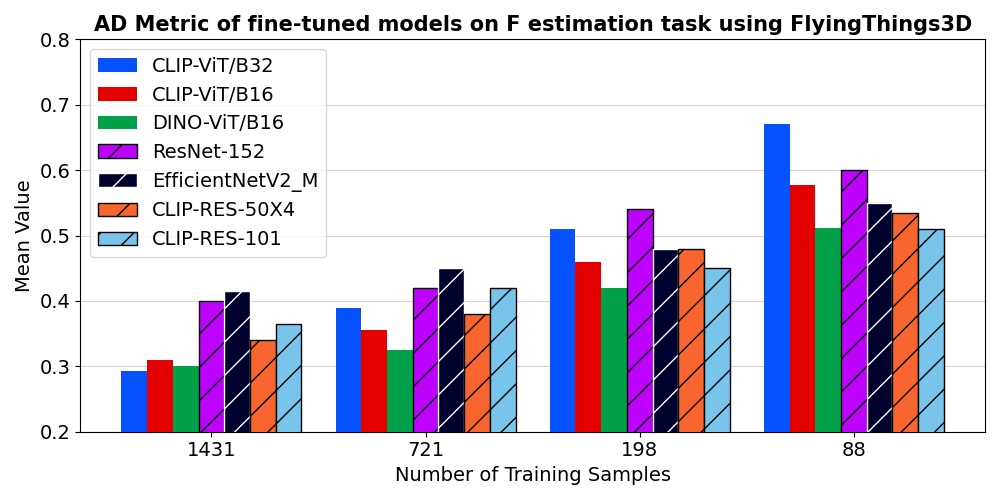}
        \textbf{(b)}
    \end{minipage}
    \caption{SED (a) and AD (b) performance of pretrained \gls{vit} and \gls{cnn}-based models in $\mathbf{F}$-matrix estimation using transfer learning with the FlyingThings3D dataset.}
    \label{fig:models_flying}
\end{figure}

In this experiment, the $\mathbf{F}$-matrix estimation performance of the pretrained \gls{vit} and \gls{cnn}-based networks was evaluated using the synthetic FlyingThings3D dataset~\cite{sceneflow}. This dataset consists of rendered synthetic sequences with different stereo baselines. Fig.~\ref{fig:models_flying} compares the \gls{vit} with the \gls{cnn}-based models' performance. Notice that the ViT-based models outperform the \gls{cnn} in scenarios with a large training size, and their performance is comparable in small-data scenarios. In addition, \glspl{vit} with smaller patch sizes outperform those with larger ones. These results further validate our findings that were made using the KITTI dataset. Notice that the overall errors are higher when using the FlyingThings3D dataset, likely due to the weak texture of images and motion artifacts. Additionally, differences in lighting conditions and occlusions might increase the errors.

\subsubsection{Cross-Domain Dataset Experiments} 
\begin{figure}
    \centering
    \begin{minipage}[t]{\linewidth}
        \centering
        \includegraphics[width=\linewidth]{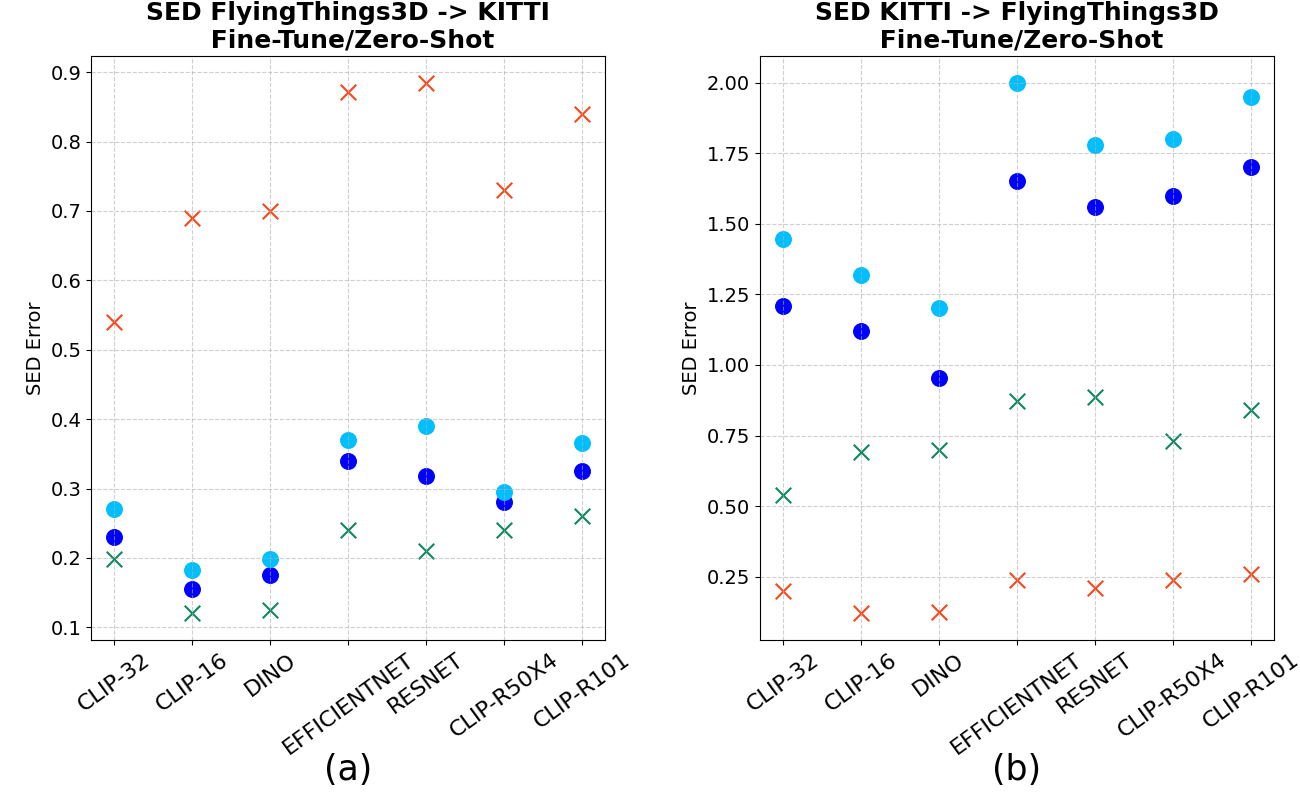}
    \end{minipage}
    \begin{minipage}[t]{\linewidth}
        \centering
        \includegraphics[width=\linewidth]{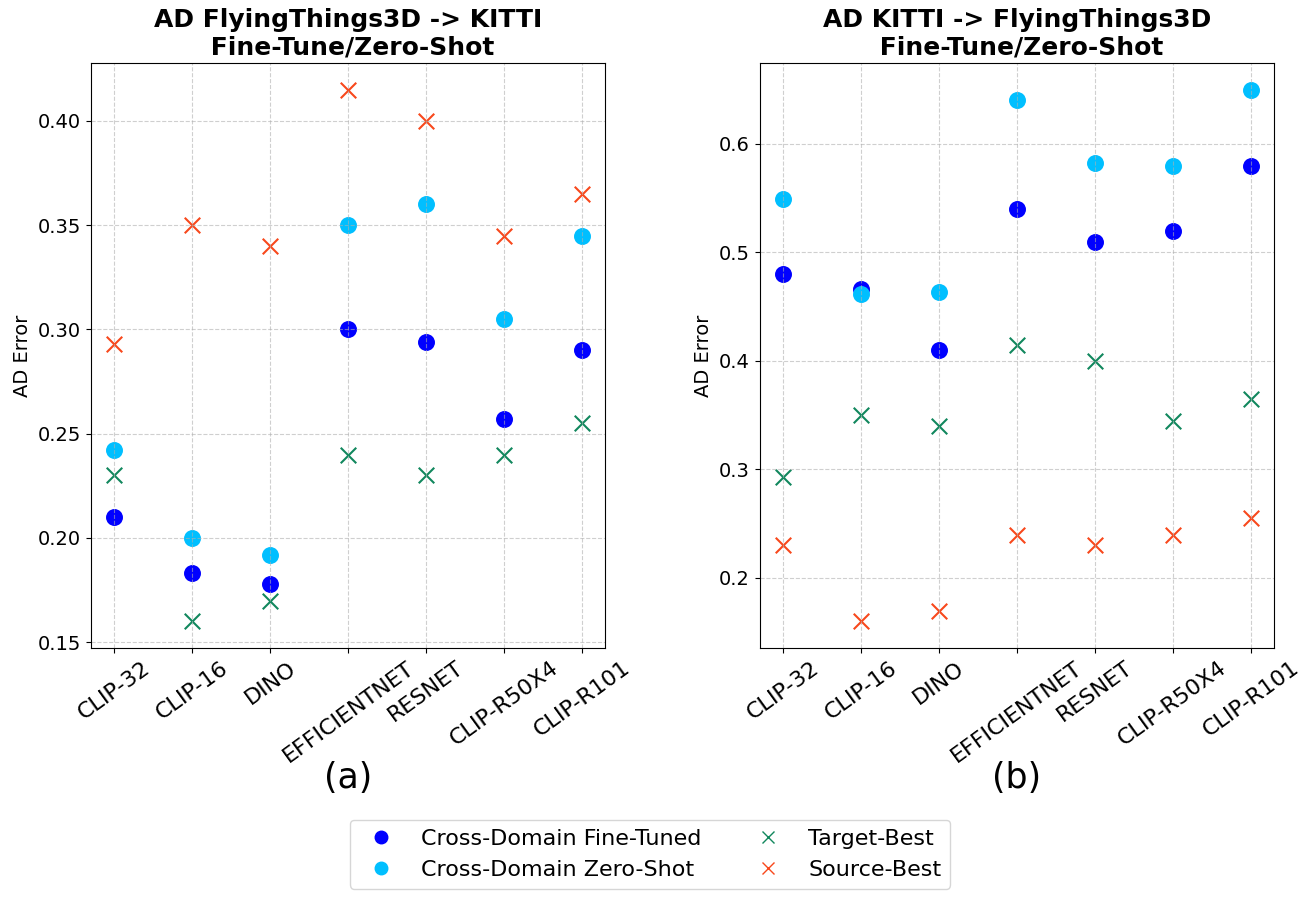}
    \end{minipage}
    \caption{Cross-domain performance evaluation. (a) Models were trained using the FlyingThings3D dataset and evaluated using the KITTI dataset. (b) Models were trained using the KITTI dataset and evaluated using the FlyingThings3D dataset. The top row is SED results, and the bottom row is AD results. 'Cross-domain fine-tuned' refers to the cross-domain evaluation of models fine-tuned with the regression head only, and 'Cross-Domain Zero-Shot' refers to models evaluated without refinement. 'Source-Best' and 'Target-Best' refer to the best-performing models trained and evaluated on the source and target datasets, respectively. We compare the results with 'Target-Best,' while 'Source-Best' serves as a reference for the evaluated models' original performance.}
    \label{fig:cross_domain}
\end{figure}

This experiment evaluated the cross-domain generalization capabilities of the considered models for $\mathbf{F}$-matrix estimation using data from different datasets. The models were trained using the source dataset and evaluated using the target dataset. Both the KITTI and FlyingThings3D datasets are used interchangeably as source and target datasets.
This experiment focuses on the best-performing models trained with the largest data size obtained in $\mathbf{F}$-matrix estimation experiments, and evaluates their cross-domain capabilities in two ways. 
First, zero-shot testing was performed by directly evaluating the model using the target dataset without any weight updates. Second, the regression module was fine-tuned for only 40 epochs using the smallest data size from the target dataset while keeping the backbone feature extractor frozen. 
Fig.~\ref{fig:cross_domain} shows the errors of the fine-tuned and zero-shot models marked as circles.
The performance of the best-performing models, trained and evaluated using the same datasets, is presented as crosses, and red crosses correspond to the source domain and green to the target domain. Green crosses mark the comparison with the cross-domain models, and red crosses mark the models' performance reference when tested using their source dataset.

Only the small performance gap between the cross-domain evaluations and the best-performing model fully trained directly using the target dataset with the largest data size demonstrates the models’ ability to generalize effectively to new domains. While both zero-shot and fine-tuned models perform well using the target dataset, the fine-tuned models slightly outperform them.
Notice that the \glspl{vit} outperform the \gls{cnn}-based models in terms of the generalization capabilities.
 This is the most prominent when comparing the CLIP-ViT-B/32 with the \gls{cnn}-based models trained on KITTI and evaluated on FlyingThings3D. While both perform similarly on the source data, the \gls{vit} outperforms them in the target domain. This suggests that although all models can successfully generalize to an unseen domain, \glspl{vit} architectures may be better suited to adapt their learned representations to diverse visual environments. Surprisingly, models trained on the synthetic FlyingThings3D dataset perform well using real complex data in zero-shot and few-shot scenarios. This can be attributed to the greater representation variability of the synthetic dataset, which allows for better generalization to KITTI data.

\subsubsection{Comparison with Alternative Approaches}
The $\mathbf{F}$-matrix estimation performance of the considered \gls{cnn} and \gls{vit} models trained using the KITTI dataset is compared with the alternative approaches from~\cite{an_end_to_end} as summarized in  Table~\ref{tab:prev}. All deep-learning-based methods are trained using similar sizes of data, except for RoMa, which was trained on very large datasets and can not be considered as the few-shot, low-data setting, which is the main scope of this work. Furthermore, it is based on dense matching principles, whereas our approach estimates the $\mathbf{F}$-matrix directly.
Table~\ref{tab:prev} shows that transfer-learning for the pretrained \gls{vit} and \gls{cnn}-based models outperform conventional methods or models trained from scratch for the considered geometric deformation estimation task.

\begin{table}[ht]
\centering
\small
\caption{Comparison of various approaches in terms of SED in a large-data scenario. For a fair comparison, the model with our training scheme is marked with *. The best result is marked in bold. Note that RoMa was trained using approximately $8$ million samples.}
\begin{tabular}{|c|c|c|c|}
\hline
    \textbf{Previous Methods} & \textbf{SED} & \textbf{Our Models} & \textbf{SED} \\ \hline
    DeepF \cite{deepf_no_cors} & 118.66 & CLIP-ViT-B/32  & 0.19 \\ \hline
    SIFT + RANSAC   & 10.88  & CLIP-ViT-B/16  & 0.17 \\ \hline
    SIFT + RANSAC * & 6.82  & Resnet-152   & 0.43 \\ \hline
    OA \cite{an_end_to_end} & 0.72  & DINO-ViT-B/16   & \textbf{0.14} \\ \hline
    RoMa \cite{roma} & \textbf{0.116} & EfficientNetV2-M  & 0.24 \\ \hline
 & & CLIP-RES-50X4&0.24\\\hline
 & & CLIP-RES-101&0.26\\\hline
\end{tabular}
\label{tab:prev}
\end{table}

\section{Ablation Study}\label{sec:short_ablation}
In this work, the \gls{vit} encoder module is optimized by evaluating four feature extraction methods using the rigid estimation task: (i) utilizing the CLS token only, (ii) using all spatial patch embeddings, (iii) averaging all patch embeddings, and (iv) applying convolutional layers with location-aware max-pooling to all patch embedding. 
Each of these methods was evaluated using different input combinations: (i)  original and transformed images, (ii) the transformed image only, (iii) original and transformed images with their element-wise product, and (iv) the element-wise product only. The results demonstrated that methods incorporating convolutional layers and spatial patch embeddings preserve local spatial details and, as a result, achieve better performance. In addition, approaches using both original and transformed images as input consistently outperformed those with transformed features only. The element-wise product had not significantly improved the performance. The detailed results are presented in~\ref{sec:long_ablation}.

\section{Conclusions and Future Work}This work compares the efficiency of pretrained \glspl{vit} and \gls{cnn}-based models as backbone encoders for fundamental matrix estimation and 2D rigid transformation prediction. The \glspl{vit}' superiority in large data scenarios and \glspl{cnn}' robustness in low-data scenarios were demonstrated via extensive performance evaluations. Experiments demonstrated that reducing the \gls{vit} capacity mitigates overfitting in data-limited settings. Moreover, different pretraining objectives play a role in the performance of the considered downstream tasks: representations learned with contrastive vision-language alignment (CLIP) or self-supervised invariance (DINO) transfer more effectively to geometric-regression tasks than those obtained from supervised classification. Moreover, the DINO-VIT/B16 model achieves comparable results with CLIP-RES-101 in the low data regime. 
The cross-domain evaluations demonstrated the \gls{vit}'s strong generalization capabilities to unseen data. Following the comparison results between CNN and ViT architectures and the comparison with RoMa  CNN-ViT architecture, our future work will explore using hybrid CNN-ViT architectures and a feature-matching approach with a transformer-decoder to predict anchor probabilities and estimate the $\mathbf{F}$-matrix directly. In addition, the dynamic estimation of the $\mathbf{F}$-matrix for a moving monocular camera or a set of cameras observing the static scene is a subject of our future work.

\bibliographystyle{ieeetr} 
\bibliography{egbib.bib}

\clearpage

\appendix
\onecolumn
\section*{\centering Appendix}

% \section*{\centering \LARGE \textbf{Appendix}}  

\addcontentsline{toc}{section}{Appendix}  
\vspace{2\baselineskip}  

\section{Losses and Evaluation Metrics}  
\label{sec:A}  

This Appendix describes the loss functions used for both the 2D rigid transformation estimation and fundamental matrix estimation tasks, as well as the evaluation metrics specific to the rigid transformation task.  

\subsection{Loss Functions}  

The training loss function in both tasks includes \gls{mse} and Huber loss, which measure the difference between the predicted and ground-truth values. The MSE loss penalizes large deviations quadratically, while the Huber loss mitigates the influence of outliers by behaving like \gls{mse} for small errors and like \gls{mae} for larger errors.
\begin{equation}
\text{MSE}(\hat{y}, y) = \frac{1}{n} \sum_{i=1}^{n} (\hat{y}_i - y_i)^2\;,
\label{eq:mse}
\end{equation}
\begin{equation}
\text{Huber loss}(\hat{y}, y) = \begin{cases} 
\frac{1}{2}(\hat{y} - y)^2 & \text{for } |\hat{y} - y| \leq \delta \\
\delta(|\hat{y} - y| - \frac{1}{2}\delta) & \text{otherwise}
\end{cases}\;,
\label{eq:huber}
\end{equation}
where \( \hat{y} \) and \( y \) denote the estimated and ground-truth values, respectively, \( n \) is the total number of samples, and \( \delta \) is a threshold parameter controlling the transition between quadratic and linear behavior.

\subsection{Evaluation Metrics For Rigid Transformation Estimation}  

The performance of the rigid transformation estimation task is evaluated using two metrics: $L2$ distance for translation error and Angle MAE for rotation error:
\begin{equation}
\text{$L2$ distance} = \frac{1}{n} \sum_{i=1}^{n} \sqrt{(\hat{x}_i - x_{\text{gt},i})^2 + (\hat{y}_i - y_{\text{gt},i})^2}\;,
\label{eq:l2}
\end{equation}
\begin{equation}
\text{Angle MAE} = \frac{1}{n} \sum_{i=1}^{n} | \hat{\theta}_i - \theta_{\text{gt}, i} |\;,
\label{eq:mae}
\end{equation}
where \( \hat{x}_i \) and \( \hat{y}_i \) are the predicted $x$ and $y$ shifts  for the \( i \)th sample, \( x_{\text{gt},i} \) and \( y_{\text{gt},i} \) are the corresponding ground-truth values, \( \hat{\theta}_i \) is the predicted rotation angle, and \( \theta_{\text{gt},i} \) is the corresponding ground-truth angle.

\section{Ablation Experiment For Feature Extraction Module} \label{sec:long_ablation}
The optimal use of the \gls{vit} based feature extractor module is obtained by evaluating a series of experiments for the 2D rigid transformation estimation task using CLIP-ViT-B/32 backbone. Various ways to utilize both the patch embedding tokens and the global CLS token were evaluated. Four approaches were evaluated: a) using only the CLS token, b) using all patch embeddings except the CLS token, c) averaging all patch embeddings, and d) applying a convolutional layer with location-aware max-pooling to all patch embeddings. 

Each of these feature extraction methods was applied separately to the input images. We further explored the most effective input combinations, which include the original and transformed images. For each of the four feature extraction methods (columns of Table~\ref{tab:ablation}), various input combinations were considered: i) using both original and transformed images as inputs, ii) only the transformed image, providing both images with their Hadamard (element-wise) product, and iii) using only the Hadamard product (rows of Table~\ref{tab:ablation}). This resulted in 16 experimental configurations summarized in Table~\ref{tab:ablation}. 

\begin{table*}[ht]
\centering
    \renewcommand{\arraystretch}{1.4} % Adjust row spacing
    \footnotesize % Reduce font size for better fit
 \caption{\gls{vit} feature extraction and input combinations in rigid estimation task. Each cell represents the Euclidean translation metric (T), in pixels, and the MAE angle metric (R), in degrees, in the format T/R. Columns represent various feature token manipulation strategies, and rows represent various input combination approaches.}
    \begin{tabular}{|l|c|c|c|c|}
        \hline
        \textbf{Method} & \textbf{Spatial patches} & \textbf{GAP} & \textbf{Convolution} & \textbf{CLS} \\
        \hline
        \textbf{Original, Transformed, Hadamard}  
        & 0.79 / 0.52  & 0.88 / 0.64  & 0.77 / 0.48  & 1.14 / 1.04  \\
        \hline
        \textbf{Original, Transformed}  
        & 0.79 / 0.56  & 0.92 / 0.68  & 0.77 / 0.52  & 1.12 / 1.32 \\
        \hline
        \textbf{Transformed}  
        & 0.81 / 2.00  & 0.92 / 2.00  & 0.83 / 1.96  & 1.36 / 2.04  \\
        \hline
        \textbf{Hadamard}  
        & 0.99 / 0.52  & 1.05 / 0.81  & 1.78 / 0.68  & 1.54 / 1.16 \\
        \hline
    \end{tabular}  
    \label{tab:ablation}
\end{table*}

\subsection{Feature Tokens Manipulation}\label{sb:columns}
Methods utilizing either convolutional layers or all spatial patches consistently achieved the lowest shift and angle estimation errors, effectively preserving local spatial information. The convolutional approach further enhances these features, maintaining their spatial integrity and improving predictive accuracy. However, applying the Hadamard product before convolution can cause feature loss due to zeroing-out effects and introduce non-linearities that disrupt essential spatial patterns. On the other hand, averaging patch embeddings or using only the CLS token led to higher errors. Global average pooling (GAP) tends to obscure fine-grained spatial details in higher layers~\cite{do_vision_transformers}, which is critical for tasks that require precise spatial localization. In contrast, though useful for global information, the CLS token fails to capture necessary local details. This emphasizes the importance of preserving local spatial information for accurate geometric transformation estimations. 

\subsection{Input Combination Approaches}\label{sb:rows}
Incorporating both original and transformed features typically yielded the best results, highlighting the necessity of providing both contexts for accurate transformation inference. The model does not perform well when relying solely on the transformed features, suggesting that unsupervised inference of transformations is ineffective without the original image for reference. The Hadamard product alone showed mixed results since it captured only some interactions and did not outperform full feature combinations, emphasizing the importance of using complete feature sets. \\

Our results demonstrate that using all available spatial information from ViT embeddings is critical for achieving low shift and angle estimation errors. Methods that compress the spatial information, such as GAP or using only the CLS token, underperformed. Additionally, using both original and transformed feature tensors consistently outperformed relying solely on the transformed features, and leveraging the Hadamard product did not provide sufficient improvement.

\section{Datasets Image Samples}
\begin{figure}
    \centering
    \begin{minipage}{0.48\linewidth}
        \centering
        \includegraphics[width=\linewidth]{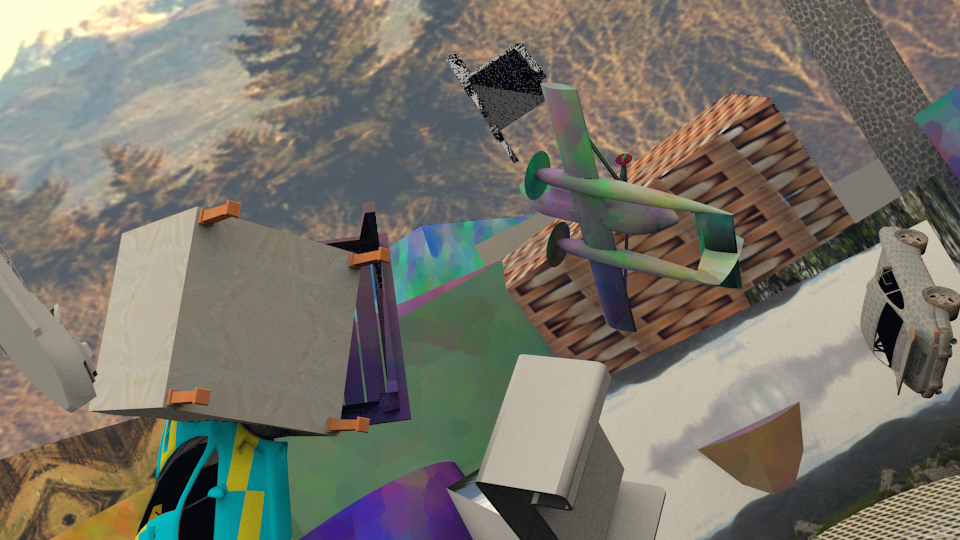}
    \end{minipage}
    \begin{minipage}{0.48\linewidth}
        \centering
        \includegraphics[width=\linewidth]{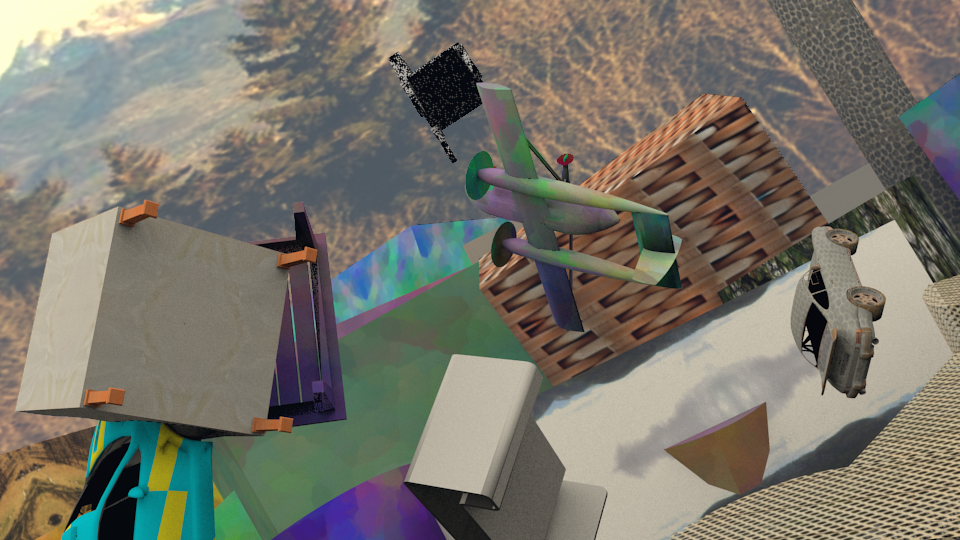}
    \end{minipage}
    
    \vspace{0.2cm} % Adjust spacing between rows

    \begin{minipage}{0.48\linewidth}
        \centering
        \includegraphics[width=\linewidth]{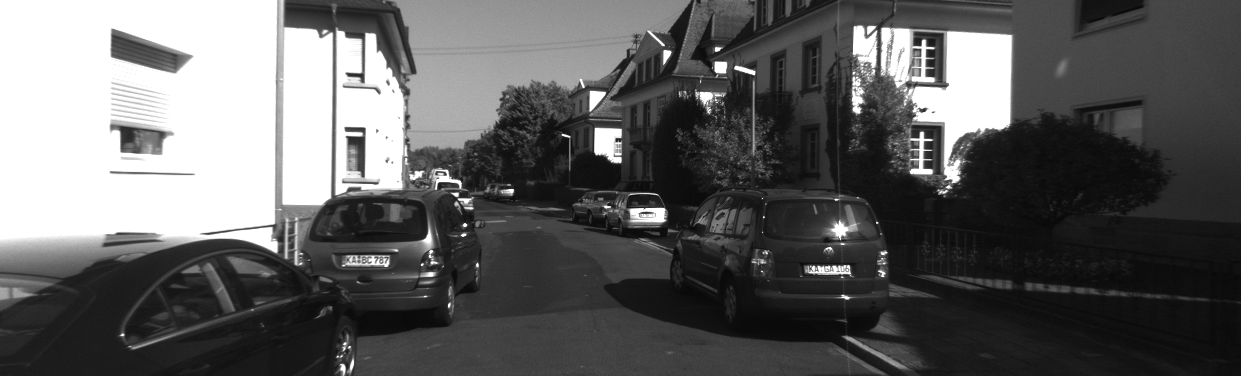}
    \end{minipage}
    \begin{minipage}{0.48\linewidth}
        \centering
        \includegraphics[width=\linewidth]{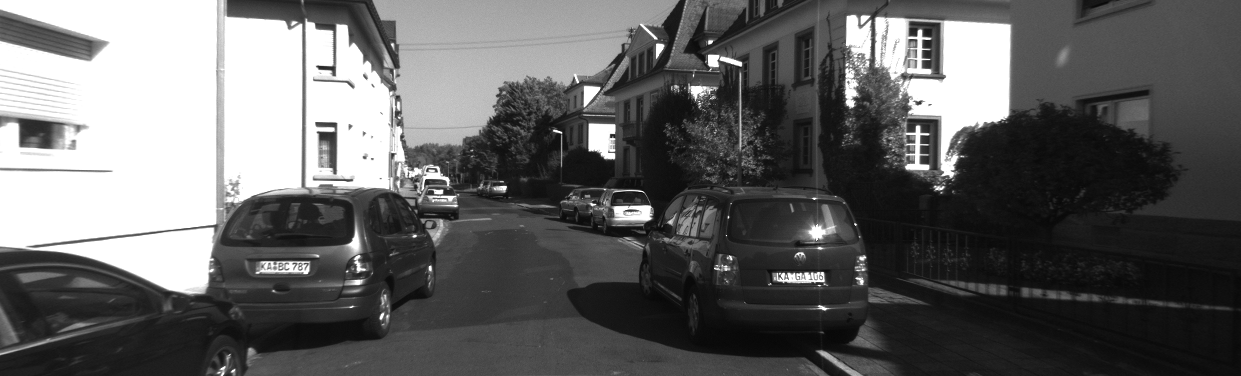}
    \end{minipage}
    \caption{Example of stereo image pairs. Top: a pair from the FlyingThings3D dataset, which consists of synthetic scenes with randomly placed objects in a 3D environment. Bottom: a pair from the KITTI dataset, which contains real-world urban driving scenes captured by a stereo camera mounted on a vehicle.} 
    \label{fig:example images}
\end{figure}

Examples of stereo image pairs from the two datasets are presented in Fig.~\ref{fig:example images} shows examples of stereo image pairs from the two datasets. These samples illustrate the differences between the two datasets. While KITTI captures real-world driving scenarios, the FlyingThings3D dataset comprises synthetic frames featuring objects moving along randomized 3D trajectories without naturalistic constraints.

\end{document}